\documentclass[12pt]{article}

\usepackage{times}
\usepackage{graphicx}
\usepackage{natbib}
\usepackage{hyperref}
\usepackage{url}

\title{Quantifying Creativity in Art Networks\footnote{This paper will be published in the sixth International Conference on Computational Creativity (ICCC) June 29-July 2nd 2015, Park City, Utah, USA. This arXiv version is an extended version of the conference paper}}
\author{\href{mailto:elgammal@cs.rutgers.edu}{Ahmed Elgammal\footnote{elgammal@cs.rutgers.edu}}     and \href{mailto:babaks@cs.rutgers.edu}{Babak Saleh\footnote{babaks@cs.rutgers.edu}}\\
\href{http://digihumanlab.rutgers.edu}{The Art and Artificial Intelligence Laboratory}  \\
Department of Computer Science\\
Rutgers University %\\ 
}

\date{}

\renewcommand{\cite}[1]{\citep{#1}}

\graphicspath{{../}}

\begin{document} 
\maketitle

\begin{abstract}
\begin{quote}

%Can the machine look at the history of art and come up with an assessment of creativity for each painting. 

Can we develop a computer algorithm that assesses the creativity of a painting given its context within art history? 
This paper proposes a novel computational framework for assessing the creativity of  creative products, such as paintings, sculptures, poetry, etc. 
We use the most common definition of creativity, which emphasizes the originality of the product and its influential value.
The proposed computational framework is based on constructing a network between creative products and using this network to infer about the originality and influence of its nodes. Through a series of transformations, we construct a Creativity Implication Network. We show that inference about creativity in this network reduces to a variant of network centrality problems which can be solved efficiently. We apply the proposed framework to the task of quantifying creativity of paintings (and sculptures). We experimented on two datasets  with over 62K paintings to illustrate the behavior of the proposed framework. We also propose a methodology for quantitatively validating the results of the proposed algorithm, which we call the ``time machine experiment''.

\end{quote}
\end{abstract}

%!TEX root =creativity_index_arxiv.tex

\section{Introduction}

The field of computational creativity is focused on giving the machine the ability to generate human-level ``creative'' products such as computer generated poetry, stories, jokes, music, art, etc., as well as creative problem solving.  An important characteristic of a creative agent is its ability to assess its creativity as well as judge other agents' creativity.  
In this paper we focus on developing a computational framework for assessing the creativity of  products, such as painting, sculpture, etc. We use the most common definition of creativity, which emphasizes the originality of the product and its influential value~\cite{PaulBarry2014Ch1}. In the next section we justify the use of this definition in contrast to other definitions.  The proposed computational framework is based on constructing a network between products and using it to infer about the originality and influence of its nodes. Through a series of transformations, we show that the problem can reduce to a variant of network centrality problems, which can be solved efficiently.

\begin{figure*}[t]
\centering
  \includegraphics[width=\linewidth]{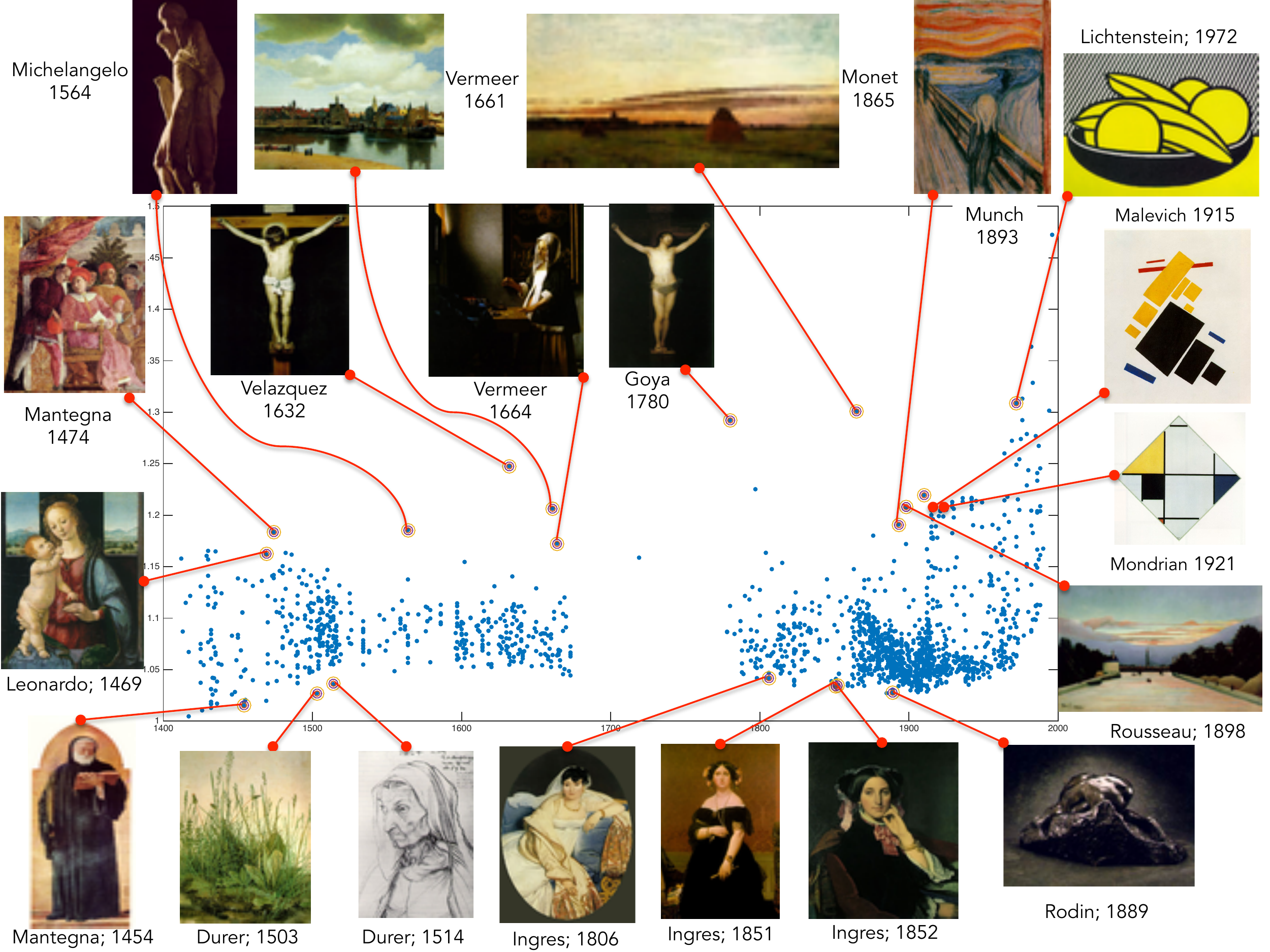}
  \caption{Creativity scores for 1710 paintings from Artchive dataset.  Each point represents a painting. The horizontal axis is the year the painting was created and the vertical axis is the creativity score (scaled). The thumbnails illustrate some of the painting that scored relatively high or low compared to their neighbors. Only artist names and dates of the paintings are shown on the graph because of limited space.  See Figure~\ref{F:artchive2} for a zoom in to the period 1850-1950}
  \label{F:artchive1}
\end{figure*}

We apply the proposed framework to the task of quantifying creativity of paintings (and sculptures).  The reader might question the feasibility, limitation, and usefulness of performing such task by a machine. Artists, art historians and critics use different concepts to describe pantings. In particular, elements of arts such as  space, texture, form, shape, color, tone and line. Artists also use principles of art including movement, unity, harmony, variety, balance, contrast, proportion, and pattern; besides brush strokes, subject matter, and other descriptive concepts~\cite{lois}. We collectively call these concepts artistic concepts.
These artistic concepts can, more or less, be quantified by today's computer vision technology. With the rapid progress in computer vision, more advanced techniques are introduced, which can be used to measure similarity between paintings with respect to a given artistic concept. Whether the state of the art is already sufficient to measure similarity in meaningful ways, or whether this will happen in the near or far future, the goal of this paper is to design a framework that can use such similarity measures to quantify our chosen definition of creativity in an objective way.  Hence, the proposed framework would provide a ready-to-use approach that can utilize any future advances in computer vision that might provide better ways for visual quantification of digitized paintings. In fact, we applied the proposed framework using state-of-the-art computer vision techniques and achieved very reasonable automatic quantification of creativity on two large datasets of paintings. Figure~\ref{F:artchive1} illustrates an example of the creativity scores obtained on  dataset containing  1710 paintings. 
%We applied the approach on two datasets, the first contains 1710 paintings and sculptures and the second one is a large-scale dataset that contains over 62K paintings. 

%What artistic concept should be used to quantify creativity? is it the use of color, the use or perspective, subject matter, brush strokes, composition, etc? Clearly, it is not possible to judge creativity based on one specific aspect. For example it was the use of perspective that characterized the creativity at certain point of art history, however it is not the same aspect for other periods. This highly suggests the need to measure creativity along different dimensions separately where each dimension reflects certain visual aspect that quantifies certain element of art. The proposes framework can be used with any artistic concept to achieve multi-dimensional creativity scoring 

 One of the fundamental issues with the problem of quantifying creativity of art is how to validate any results that the algorithm can obtain. Even if art historians would agree on a list of highly original and influential paintings that can be used for validation, any algorithm that aims at assigning creativity scores will encounter three major limitations: I) Closed-world limitation: The algorithm is only limited to the set of paintings it analyzed. It is a closed world for the algorithm where this set is every thing it has seen about art history. The number of images of paintings available in the public domain is just a small fraction of what are in museums and private collections. II) Artistic concept quantification limitation: the algorithm is limited by what it sees, in terms of the ability of the underlying computer vision methods to encode the important elements and principles of art that relates to judging creativity.
 III) Parameter setting: the results will depend on the setting of the parameters, where each setting would mean a different way to assign creativity scores with different interpretation and different criteria. 
However, these limitations should not stop us from developing and testing algorithms to quantify creativity. The first two limitations are bound to disappear in the future, with more and more paintings  being digitized, as well as with the continuing advances in computer vision and machine learning. The third limitation should be thought of as an advantage, since the different settings mean a rich ability of the algorithm to assign creativity scores based on different criteria. For the purpose of validation, we propose a methodology for validating the results of the algorithm through what we denote as ``time machine experiments'', which provides evidence of the correctness of the algorithm.

Having discussed the feasibility and limitations, let us discuss the value of using any computational framework to assess creativity in art. For a detailed discussion about the implications of using computational methods in the domain of aesthetic-judgment-related tasks, we refer the reader to~\cite{spratt2014computational}. Our goal is not to replace art historians' or artists' role in judging creativity of art products. Providing a computational tool that can process millions of artworks to provide objective similarity measures and assessments of creativity, given certain visual criteria can be useful in the age of digital humanities. From a computational creativity point of view, evaluating the framework on digitized art data provides an excellent way to optimize and validate the framework, since art history provides us with suggestions about what is considered creative and what might be less creative. In this work we did not use any such hints in achieving the creativity scores, since the whole process is unsupervised, i.e., the approach does not use any creativity, genre, or style labels. However we can use evidence from art history to judge whether the results make sense or not.   Validating the framework on digitized art data makes it possible to be used on other products where no such knowledge is available, for example to validate computer-generated creative products.

\section{On the Notion of Creativity}

There is a historically long and ongoing debate on how to define creativity. In this section we give a brief description of some of these definitions that directly relate to the notion we will use in the proposed computational framework. Therefore, this section is by no means intended  to  serve as a comprehensive overview of the subject. We refer readers to~\cite{taylor1988various,PaulBarry2014} for comprehensive overviews of the different definitions of creativity. 

 We can describe a person (e.g. artist, poet), a product (painting, poem), or the mental process as being creative~\cite{taylor1988various,PaulBarry2014Ch1}. Among the various definitions of creativity it seems that there is a convergence to two main conditions for a product to be called ``creative''. That product must be novel, compared to prior work, and also has to be of value or influential~\cite{PaulBarry2014Ch1}. These criteria resonate with Kant's definition of artistic genius, which emphasizes two conditions  ``originality'' and being  ``exemplary'' \footnote{ Among four criteria for artistic genius suggested by Kant, two describe the characteristic of a creative product ``That genius 1) is a talent for producing that for which no determinate rule can be given, not a predisposition of skill for that which can be learned in accordance with some rule, consequently that originality must be it's primary characteristic. 2) that since there can also be original nonsense, its products must at the same time be models, i.e., exemplary, hence, while not themselves the result of imitation, they must yet serve others in that way, i.e., as a standard or rule for judging.'' \cite{guyer2000critique}-p186}.  Psychologists would not totally agree with this definition since they favor associating creativity with the mental process that generates the product~\cite{taylor1988various,Nanay2014Ch}. However associating creativity with products makes it possible to argue in favor of  ``Computational Creativity'', since otherwise, any computer product would be an output of an algorithmic process and not a result of a creative process. Hence, in this paper we stick to quantifying the creativity of products instead of the mental process that create the product. 
 
Boden suggested a distinction between two notions of creativity: psychological creativity (P-creativity), which assesses novelty of ideas with respect to its creator,  and historical creativity (H-creativity), which assesses novelty with respect to the whole human history~\cite{boden1990creative}.  It follows that P-creativity is a necessary but not sufficient condition for H-creativity, while H-creativity implies P-creativity~\cite{boden1990creative,Nanay2014Ch}. This distinction is related to the subjective (related to person) vs. objective creativity (related to the product) suggested by Jarvie~\cite{jarvie1986rationality}. In this paper our definition of creativity is aligned with objective/H-creativity, since we mainly quantify creativity within a historical context.

\section{Computational Framework}
%Creativity definition
%

According to the discussion in the previous section, a creative product must be {\em original}, compared to prior work, and valuable ({\em influential}) moving forward.
Let us construct a network of creative products and use it to assign a creativity score  to each product in the network according to the aforementioned criteria. In this section, for simplicity and without loss of generality, we describe the approach based on a network of paintings, however the framework is applicable to other art or literature forms.

\begin{figure*}[t]
\center
  \includegraphics[height=2.2in,width=\linewidth]{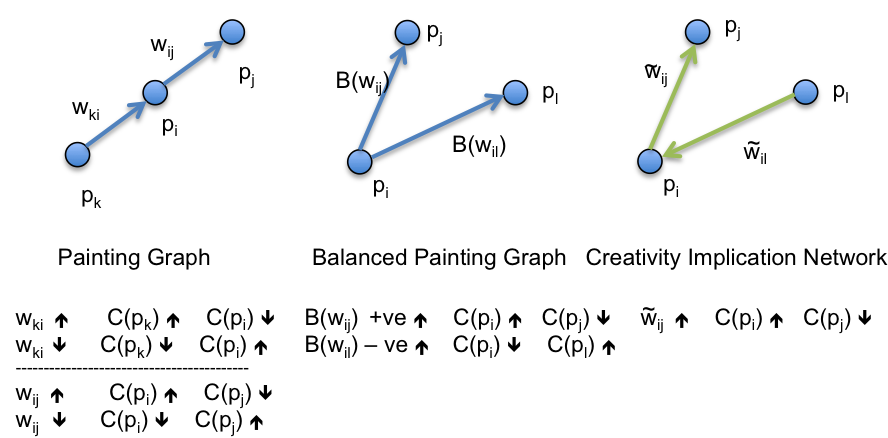}
  \caption{Illustration of the construction of the Creativity Implication Network: blue arrows indicate temporal relation and orange arrows indicate reverse creativity implication (converse).}
  \label{F:CIN}
\end{figure*}

%Painting graph
%\medskip
%\noindent{\bf Constructing a Painting Graph:}

\subsection{Constructing a Painting Graph}
Let us denote by $P=\{p_i, i=1\cdots N \}$ a set of paintings. The goal is to assign a creativity score for each painting, denoted by $C(p_i)$ for painting $p_i$ . Every painting comes with a time label indicating the date it was created, denoted by $t(p_i)$.  
We create a directed graph where each vertex corresponds to a painting.  A directed edge (arc) connects painting $p_i$ to $p_j$  if  $p_i$ was created before $p_j$.  Each directed edge is assigned a positive weight (we will discuss later where the weights come from), we denote the weight of edge $(p_i, p_j)$ by $w_{ij}$. We denote by $W_{ij}$ the adjacency matrix of the painting graph, where $W_{ij} = w_{ij} $ if there is an edge from $p_i$ to $p_j$ and 0 otherwise.  Note that according to this definition, a painting is not connected to itself, i.e., $w_{ii}=0, i=1\cdots N$.  By construction,  $w_{ij} > 0 \rightarrow w_{ji}=0$, i.e., the graph is anti-symmetric. 

%Similarity function based on a specific visual aspect
%
%Weights reflect similarity..
%Wij

To assign the weights we assume that there is a similarity function that takes two paintings and produces a positive scalar measure of affinity between them (higher value indicates higher similarity). We denote such a function by $S(\cdot,\cdot)$ and, therefore, %$w_{ij}=f(p_i,p_j)$. 
\begin{eqnarray*}
w_{ij}=
\left\{
\begin{array}{ll}
  S(p_i,p_j) &  \hbox{if}\;\; t(p_i) < t(p_j).     \\
  0 &  \hbox{otherwise.}    
\end{array}
\right.
\end{eqnarray*}
Since there are multiple possible visual aspects that can be used to measure similarity, we denote such a function by $S^a(\cdot,\cdot)$ where the superscript $a$ indicates the visual aspect that is used to measure the similarity (color, subject matter, brush stroke, etc.)   This implies that we can construct multiple graphs, one for each similarity function. We denote the corresponding adjacency matrix by $W^a$, and the induced creativity score by $C^a$, which measure the creativity along the dimension of visual aspect $a$. In the rest of this section, for the sake of simplicity, we will assume one similarity function and drop the superscript. Details about the similarity function will be explained in the next section.

%Moving from creativity definition to processing the graph
%
%Higher w
%
%Lower w
%

%\medskip
%\noindent{\bf Creativity Propagation:}

\subsection{Creativity Propagation}
Giving the constructed painting graph, how can we propagate the creativity in such a network? To answer this question we need to understand the implication of the weight of the directed edge connecting two nodes on their creativity scores.  Let us assume that initially we assign equal creativity indices to all nodes. Consider painting $p_i$ and consider an incoming edge from a prior painting $p_k$. 
 A high weight on that edge ($w_{ki}$) indicates a high similarity between $p_i$ and $p_k$, which indicates that $p_i$ is not novel, implying that we should lower the creativity score of $p_i$ (since $p_i$ is subsequent to $p_k$ and similar to it) and increase the creativity score of $p_k$. In contrast, a low weight implies that $p_i$ is novel and hence creative compared to $p_k$, therefore we need to increase the creativity score of $p_i$ and decreases that of $p_k$. 
 
 Let us now consider the outgoing edges from $p_i$. According to our notion of creativity, for $p_i$ to be creative it is not enough to be novel, it has to be influential as well (some others have to imitate it). This indicates that a high weight, $w_{ij}$, between $p_i$ and a subsequent painting $p_j$ implies that we should increase the creativity score of $p_i$ and decrease that of $p_j$. In contrast, a lower weight implies that $p_i$ is not influential on $p_j$, and hence we should decrease the score for $p_i$ and increase it for $p_j$. These four cases are illustrated in Figure~\ref{F:CIN}. A careful look reveals that the two cases for the incoming edges and those for the outgoing edges are in fact the same. {\em A higher weight implies the prior node is more influential and the subsequent node is less creative, and a lower weight implies the prior node is less influential and the subsequent node is more creative.}

%Balancing function
%

%\medskip
%\noindent{\bf Creativity Implication Network:}

\subsection{Creativity Implication Network}
Before converting this intuition to a computational approach, we need to define what is considered high and low for weights. We introduce a balancing function on the graph. Let $m(i)$ denote a balancing value for node $i$, where for the edges connected to that node a weight above  $m(i)$ is considered high and below that value is considered low. We define a balancing function as a linear function on the weights connecting to each node in the form 
\[B_i(w) =
\left\{
\begin{array}{ll}
  w-m(i) &   \hbox{if}\;\; w>0 . \\
  0 &  \hbox{otherwise.}   
\end{array}
\right.
\]
%We denote the weights after balancing by {\em balanced weights}. 
We can think of different forms of balancing functions that can be used. Also there are different ways to set the parameter $m(i)$ with different implications, which we will discuss in the next section. This form of balancing function basically converts weights lower than $m(i)$ to negative values. The more negative the weight of an edge the more creative the subsequent node and the less influential the prior node. The more positive the weight of an edge the less creative the subsequent node and the more influential the prior node.

%Reversing the edges
%

The introduction of the negative weights in the graph, despite providing a solution to  represent low weights, is problematic when propagating the creativity scores. The intuition is,  a negative edge between $p_i$ and $p_j$ is equivalent to a positive edge between $p_j$ and $p_i$. This directly suggests that we should reverse all negative edges and negate their values. Notice that the original graph construction guarantees that an edge between $p_i$ and $p_j$ implies no edge between $p_j$ and $p_i$, therefore there is no problem with edge reversal. This process results in what we call {\em ``Creativity Implication Network''}. We denote the weights of that graph by $\tilde{w}_{ij}$ and its adjacency matrix by $\widetilde{W}$. This process can be described mathematically as
\begin{eqnarray*}
  B(w_{ij}) > 0 &  \rightarrow  & \tilde{w}_{ij} = B(w_{ij})  \\
  B(w_{ij}) =  0 & \rightarrow  &Ê \tilde{w}_{ij} = 0    \\
  B(w_{ij}) < 0 &  \rightarrow  &   \tilde{w}_{ji} =  - B(w_{ij})
\end{eqnarray*}
The  Creativity Implication Network has one simple rule that relates its weights to creativity propagation: {\em the higher the weight of an edge between two nodes, the less creative the subsequent node and the more creative the prior node.} Note that the direction of the edges in this graph is no longer related to the temporal relation between its nodes, instead it is directly inverse to the way creativity scores  should propagate from one painting to another.  Notice that the weights of this graph are non-negative.

%\medskip
%\noindent{\bf Computing Creativity Scores:}

\subsection{Computing Creativity Scores}
Given the construction of the Creativity Implication Network, we are now ready to define a recursive formula for assigning creativity scores. We will show that the construction of the Creativity Implication Network reduces the problem of computing the creativity scores to a traditional network centrality problem.  The algorithm will maintain creativity scores that sum up to one, i.e., the creativity scores form a probability distribution over all the paintings in our set.  Given an initial equal creativity scores, the creativity score of node $p_i$ should be updated as
\begin{equation}
    C(p_i)= \frac{(1-\alpha)}{N} + \alpha \sum_j  \tilde{w}_{ij} \frac{C(p_j)}{N(p_j)},
    \label{E:creativity_recursive}
\end{equation}
where $ 0 \leq \alpha \leq 1 $ and $N(p_j)= \sum_k \tilde{w}_{kj}$.
In this formula, the creativity of  node $p_i$ is computed from aggregating a fraction $\alpha$ of  the creativity scores from its outgoing edges weighted by the adjusted weights $\tilde{w}_{ij}$. The constant term ${(1-\alpha)}/{N}$ reflects the chance that similarity between two paintings might not necessarily indicate that the subsequent one is influenced by the prior one. For example, two paintings might be similar simply because they follow a certain style or art movement. The factor $1-\alpha$ reflects the  probability of this chance. The normalization term $N(p_j)$ for node $j$ is the sum of its incoming weights, which means that the contribution of node $p_j$ is split among all its incoming nodes based on the weights, and hence, $p_i$ will collect only a fraction ${\tilde{w}_{ij}}/{\sum_k \tilde{w}_{kj}} $ of the creativity score of $p_j$. 

The recursive formula in Eq~\ref{E:creativity_recursive} can be written in a matrix form as
\begin{equation}
   C = \frac{(1-\alpha)}{N} \mathbf{1} + \alpha \widetilde{\widetilde{W}} C,
   \label{E:creativity_matrix}
\end{equation}
where $\widetilde{\widetilde{W}}$ is a column stochastic matrix defined as $\widetilde{\widetilde{W}}_{ij} = {\tilde{w}_{ij}}/{\sum_k \tilde{w}_{kj}} $, and $\mathbf{1}$ is a vector of ones of the same size as $C$. It is easy to see that since $\widetilde{\widetilde{W}}$, $C$, and $ \frac{1}{N}\mathbf{1} $ are all column stochastic, the resulting scores will always sum up to one.  The creativity scores can be obtained by iterating over Eq~\ref{E:creativity_matrix} until conversion. Also a closed-form solution for the case where $\alpha \neq 1$ can be obtained as
\begin{equation}
  C^*=\frac{(1-\alpha)}{N} (I-\alpha  \widetilde{\widetilde{W}} )^{-1} \mathbf{1}.
  \label{E:closed-form}
\end{equation}

A reader who is familiar with social network analysis literature might directly see the relation between this formulation and some traditional network centrality algorithms. 
Eq~\ref{E:creativity_matrix} represents a random walk in a Markov chain. Setting $\alpha=1$, the formula in Eq~\ref{E:creativity_matrix} becomes a weighted variant to eigenvector centrality~\cite{borgatti2006graph}, where a solution can be obtained by the right eigenvector corresponding to the largest eigenvalue of $ \widetilde{\widetilde{W}} $. The formulation in Eq~\ref{E:creativity_matrix} is also a weighted variant of Hubbell's  centrality~\cite{hubbell1965input}. Finally the formulation can be seen as an inverted weighted variant of the Page Rank algorithm~\cite{brin1998anatomy}. Notice that this reduction to traditional network centrality formulations was only possible because of the way the Creativity Implication Network was constructed.

\subsection{Originality vs. Influence}
The formulation above sums up the two criteria of creativity, being original and being influential. We can modify the formulation to make it possible to give more emphasis to either of these two aspects when computing the creativity scores. For example it might be desirable to  emphasize novel works even though they are not influential, or the other way around. Recall that the direction of the edges in Creativity Implication Network are no longer related to the temporal relation between the nodes. 
We can label (color) the edges in the network such that each outgoing edge $e(p_i,p_j)$ from a given node $p_i$ is either labeled as a subsequent edge or a prior edge depending on the temporal relation between $p_i$ and $p_j$. This can be achieved by 
defining two disjoint subsets of the edges in the networks 
\begin{eqnarray*} 
E^{\hbox{prior}} &=& \{e(p_i,p_j) : t(p_j) < t(p_i)  \} \\
E^{\hbox{subseq}} &=& \{e(p_i,p_j) : t(p_j) \geq t(p_i)  \} 
\end{eqnarray*}
This results in two adjacency matrices, denoted by $\widetilde{W}^{{p}}$ and $\widetilde{W}^{{s}}$ such that $\widetilde{W} = \widetilde{W}^{{p}} + \widetilde{W}^{{s}}$, where the superscripts $p$ and $s$ denote the prior and subsequent edges respectively. Now Eq~\ref{E:creativity_recursive} can be rewritten as 
{\small
\begin{eqnarray}
    C(p_i)&=& \frac{(1-\alpha)}{N} +  \\ \nonumber
     && \alpha [ \beta \sum_j  \tilde{w}^{{p}}_{ij} \frac{C(p_j)}{N^p(p_j) }  + (1-\beta) \sum_j  \tilde{w}^{{s}}_{ij} \frac{C(p_j)}{N^s(p_j)} ],
   \label{E:creativity_recursive2}
\end{eqnarray}
} where $N^p(p_j)= \sum_k \tilde{w}^p_{kj}$ and $N^s(p_j)= \sum_k \tilde{w}^s_{kj}$. The first summation collects the creativity scores stemming from prior nodes, i.e., encodes the originality part of the score, while the second summation collects creativity scores stemming from subsequent nodes, i.e, encodes influence. We introduced a parameter $ 0 \leq \beta \leq 1 $ to control the effect of the two criteria on the result. The modified formulation above can be written as
\begin{equation}
    C = \frac{(1-\alpha)}{N} \mathbf{1} + \alpha [\beta \widetilde{\widetilde{W^p}} C + (1-\beta) \widetilde{\widetilde{W^s}} C ] ,
   \label{E:creativity_matrix2}
\end{equation}  
where $\widetilde{\widetilde{W^{p}}}$ and $\widetilde{\widetilde{W^s}}$ are the column stochastic adjacency matrices resulting from normalizing the columns of  $\widetilde{W}^{{p}}$ and $\widetilde{W}^{{s}}$  respectively. It is obvious that the closed-form solution in Eq~\ref{E:closed-form} is applicable to this modified formulation where $\widetilde{\widetilde{W}}$ is defined as $\widetilde{\widetilde{W}} = \beta \widetilde{\widetilde{W^p}}  + (1-\beta) \widetilde{\widetilde{W^s}}  $.

\section{Creativity Network for Art}
%In the previous section we introduced the computational framework in general terms for any network of creative products. 
In this section we explain how the framework can be realized for the particular case of visual art. %This will be explained by giving exact definition for the similarity function, the balancing functions, as well as discussing the effect of the different parameter of the framework.

{\em Visual Likelihood:}
For each painting we can use computer vision techniques to obtain different feature representations for its image, each encoding a specific visual aspect(s) related to the elements and principles of arts. We denote such features by $f^a_i$ for painting $p_i$, where $a$ denotes the visual aspect that the feature quantifies. 
We define the similarity between painting $p_i$ and $p_j$, as the likelihood that painting $p_j$ is coming from a probability model defined by painting $p_i$. In particular, we assume a Gaussian probability density model for painting $p_i$, i.e.,
\[ \small
  S^a(p_j,p_i) = Pr(p_j | p_i, a) = \mathcal{N}(\cdot; f^a_i,\sigma^a I).
\]
It is important to limit the connections coming to a given painting. By construction, any painting will be connected to all prior paintings in the graph.  This makes the graph highly biased since modern paintings will have extensive incoming connections and early paintings will have extensive outgoing connections. Therefore we limit the incoming connections to any node to at most the top $K$ edges (the  $K$ most similar prior paintings).

{\em Temporal Prior:} It might be desirable to add a temporal prior on the connections.  If a painting in the nineteenth century resembles a painting from the fourteenth century, we shouldn't necessarily penalize that as low creativity. This is because certain styles are always reinventions of older styles, for example neoclassicism and renaissance. Therefore, these similarities between styles across  distant time periods should not be considered as low creativity. Therefore, we can add a temporal prior to the likelihood as
\[ \small
  S^a(p_j,p_i) = Pr^v(p_j | p_i, a) \cdot Pr^t(p_j | p_i),
\]
where the second probability is a temporal likelihood (what is the likelihood that $p_j$ is influenced $p_i$ given their dates) and the first is the visual likelihood. There are different ways to define such a temporal likelihood. The simplest way is a temporal window function, i.e., $Pr^t(p_j | p_i) = 1 $ if $p_i$ is within K temporal neighbors prior to $p_j$ and 0 otherwise\footnote{Alternatively,  a Gaussian density can be use, $Pr^t(p_j | p_i) = \exp(- [t(p_i)-t(p_j)]^2 / \sigma_t^2) $. However, adding such temporal Gaussian would complicate the algorithm since it will not be easy to estimate a suitable $\sigma_t$, specially the graph can have non-uniform density over the time line.}.

{\em Balancing Function:}
There are different choices for the balancing function $B(w)$, as well as the parameter for that function. We mainly used a linear function for that purpose. The parameter $m$ can be set globally over the whole graph, or locally for each time period. A global $m$ can be set as the p-percentile of the weights of the graph, which is p-percentile of all the pairwise likelihoods. This directly means that p\% of the edges of the graph will be reversed when constructing the Creativity Implication Graph. One disadvantage of a global balancing function is that different time periods have different distributions of weights. This suggests using a local-in-time balancing function. To achieve that we compute $m_i$ for each node as p\% of the weight distribution based on its temporal neighborhood.

%Creativity flow network - better name?
%Creativity implication network
%
%
%Objective function and justification
%
%Iterative solution
%Closed form solution
%
%graph creation
%
%Likelihood computation 
%Historical prior
%Adaptive kernel
%Choice of balancing function
%
%Relate. To network literature

\section{Experiments and Results}

\subsection{Datasets and Visual Features}
{\em Artchive:}
This dataset was previously used for style classification and influence discovery~\cite{Saleh2014}.  It contains a total of 1710 images of art works (paintings and sculptures) by 66 artists, from 13 different styles ranging from AD 1412 to 1996, chosen from Mark Harden's Artchive database of fine-art~\cite{artchive}.  The majority of these images are of the full work, while a few are details of the work. 
%These include, with no specific order, Expressionism (10 artists), Impressionism (10), Renaissance (12), Romanticism (5), Cubism (4), Baroque (5), Pop (4), Abstract Contemporary (7), Surrealism (2), American Modernism (2), Post-Impressionism (3), Symbolism (1), and Neoclassical (1).

{\em Wikiart.org:}
We used the publicly available dataset of \textit{``Wikiart paintings''}\footnote{http://www.wikiart.org/}; which, to the best of our knowledge, is the largest online public collection of artworks. This collection has images of 81,449 fine-art paintings and sculptures from 1,119 artist spanning from 1400 till after 2000. These paintings are from 27 different styles (Abstract, Byzantine, Baroque, etc.) and 45 different genres (Interior, Landscape, Portrait, etc.). 
%A part of this dataset was recently used for style classification~\cite{Bar2014}. 
We pruned the dataset to 62,254 western paintings by removing genres and mediums that are not suitable for the analysis such as sculpture, graffiti, mosaic, installation, performance, photos, etc. 

For both datasets the time annotation is mainly the year. Therefore, it is not possible to tell which is prior between any pair of paintings with the same year of creation. Therefore no edge is added between their corresponding nodes. 

%\subsection{Visual Features}
We experimented with different state-of-the-art feature representations. In particular, the results shown here are using Classeme features~\cite{Torresani2010}. These features were shown  to outperform other state-of-the-art features for the task of style classification~\cite{Saleh2014}.  These features (2659 dimensions) provide semantic-level representation of images, by encoding the presence of a set of basic-level object categories (e.g. horse, cross, etc.), which captures the subject matter of the painting. Some of the low-level features used to learn the Classeme features also capture the composition of the scene. We also experimented with GIST features, which mainly encode scene decomposition along several perceptual dimensions (naturalness, openness, roughness, expansion, ruggedness)~\cite{GIST}. GIST features are widely used in the computer vision literature for scene classification.

%These features provide semantic-level encoding of images, and hence can be thought of as an embedding of the painting to a semantic space, which captures the subject matter of the painting. These features are designed to encode the presence of a set of basic-level object categories as following: a list of entry-level categories (e.g. horse, cross, etc.) is used for downloading a large collection of images from the web. The categories and the images are general and not related to domain of art.  For each image a set of low-level visual features are extracted and one classifier is learned for each category. For a given test image, these classifiers are applied on the image and their responses (confidences) make the feature vector. We followed the implementation of~\cite{classemes} and for each image extracted a 2659 dimensional real-valued Classeme feature vector. 

\subsection{Example Results}
\label{S:Results}

We show qualitative and quantitative experimental results of the framework applied to the aforementioned datasets. As mentioned in the introduction, any result has to be evaluated given the set of paintings available to the algorithm and the capabilities of the visual features used. Given that the visual features used are mainly capturing subject matter and scene composition, sensible creativity scores are expected to reflect these concept. A low creativity score does not mean that the work is not creative in general, it just means that the algorithm does not see it  creative with respect to its encoding of subject matter and composition.

Figures~\ref{F:artchive1} shows the creativity scores computed for the Artchive dataset\footnote{For Figure~\ref{F:artchive1} a temporal prior was used.  We set K=500, $\alpha$=0.15.}. In this figure and all following figures we plot the scores vs. the year of the painting. The figures visualize some of the paintings that obtained high scores, as well as some with low scores (the scores in the plots are scaled). We randomly sampled points with low scores for visualization. A close look at the paintings that scored low (bottom of each plot) reveals the presence of typical subject matter, or in some cases the image presents an unclear view of a sculpture (e.g. Rodin 1889 sculpture in the bottom right of Figure~\ref{F:artchive1}).

There are several interesting paintings that achieved high creativity scores. For example, the scream by Edvard Munch's (1893) scored very high relative to other  paintings in that period (see Figure~\ref{F:artchive1}). This painting is considered as the second iconic figure after Leonardo's Mona Lisa in the history of art,  and it is  known to be the  most-reproduced painting in the twentieth century~\cite{JohnsonBook}. It is also one of the most outstanding expressionist paintings.

\begin{figure*}[tp]
\centering
 \includegraphics[width=0.85\linewidth]{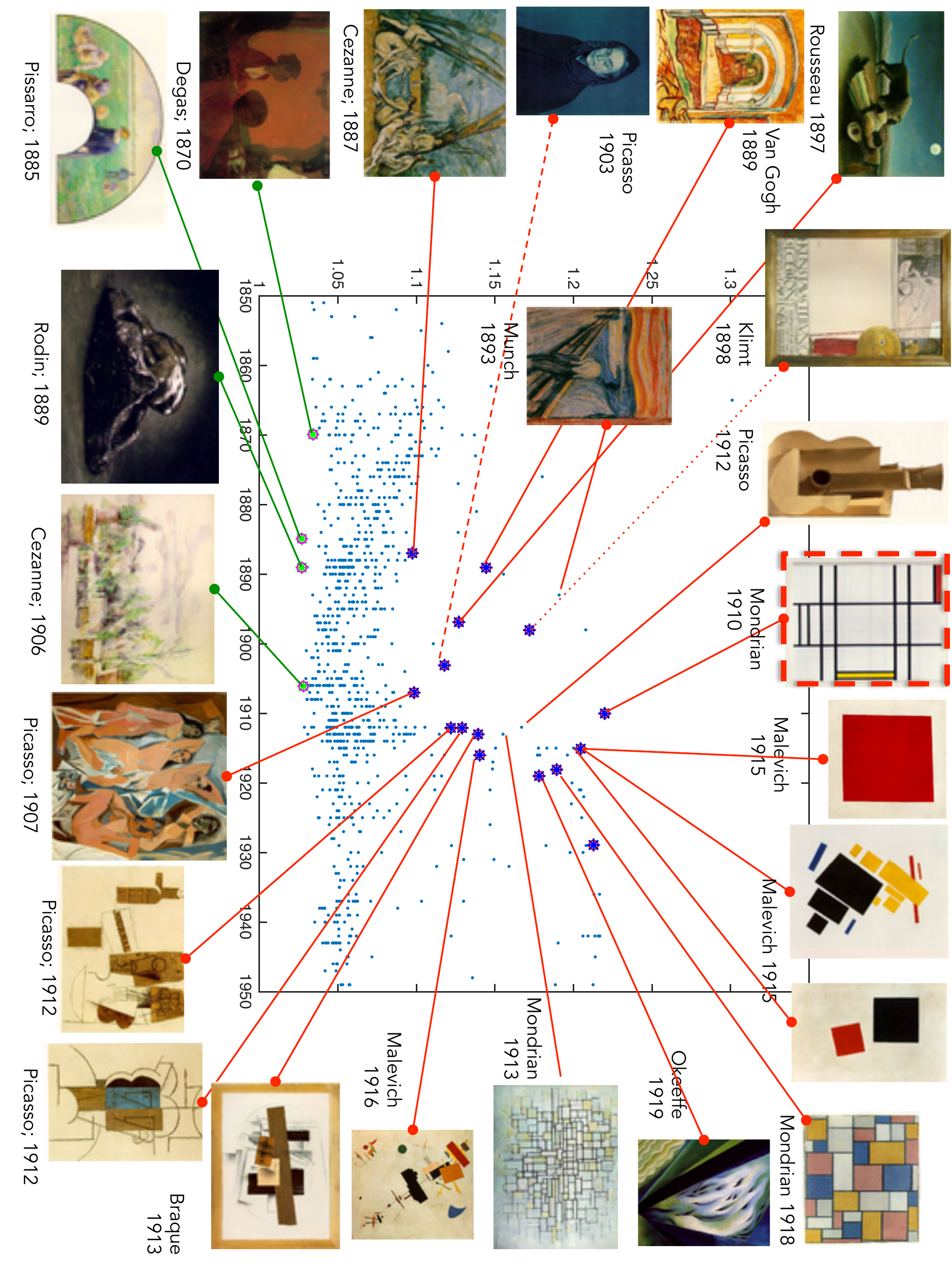}
  \caption{Zoom in to the period of 1850-1950 from Figure~\ref{F:artchive1}. Each point represents a painting. The horizontal axis is the year the painting was created and the vertical axis is the creativity score (scaled).  Only artist names and dates of the paintings are shown on the graph because of limited space. The red-dotted-framed painting by Piet Mondrain scored very high because it was wrongly dated in the dataset to 1910 instead of 1936. See Section~\ref{S:Results} for a detailed explanation.}
  \label{F:artchive2}
\end{figure*}

Figure~\ref{F:artchive2} shows a zoom-in plot to the period between 1850-1950, which is very dense in the graph of Figure~\ref{F:artchive1}. We can see that Picasso's La Celestina (1903) scored the highest among his blue-period paintings. Picasso's Ladies of Avignon (1907) sticks out as high in creativity (obtained the highest score between 1904-1911).  Art historians indicate that the flat picture plane and the application of primitivism in this painting made it an innovative work of art, which lead to Picasso's cubism~\cite{Cooper1971}.  We can notice a sharp increase in creativity scores at 1912, dominated by cubism work, with Picasso's Maquette for Guitar (1912) is the highest scoring in that surge.  The up trend in creativity scores continues with several of Kasimir Malevich's first Suprematism paintings in 1915 topping the scores. This includes Malevich's Red square (1915),  Airplane Flying (1915) and Black and Red square (1915) with almost identical scores at the top of this group, followed by  Suprematist Construction (1915),  Two-dimensional Self Portrait (1915), and Supermatist Composition (1915) - See Figure~\ref{F:artchive2} (the thumbnails of the last three paintings are not shown in the figure). Malevich's 1915's Black Square was not included in the analyzed collection that is used for this plot. However, his 1929's version of the Black Square was part of the collection and scored as high  (see the blue star around the year 1929 in Figure~\ref{F:artchive2}). The majority of the top-scoring paintings between 1916 and 1945 were by Piet Mondrian  and Georgia O'Keeffee.

One of the interesting findings is the ability of our algorithm to point out wrong annotations in the dataset. For example, one of the highest scoring paintings around 1910 was a painting by Piet Mondrian called `` Composition en blanc, rouge et jaune,''
%\footnote{Los Angeles County Museum of Art}
 (see the red-dotted-framed painting in Figure~\ref{F:artchive2}). By examining this painting, we found that the correct date for it is around 1936 and it was mistakenly annotated in the Artchive dataset as 1910\footnote{The wrong annotation is in the Artchive CD obtained in 2010. The current online version of Artchive has corrected annotation for this painting}. Modrain did not start to paint in this grid-based (Tableau) style untill around 1920. So it is no surprise that wrongly dating one of Mondrain's tableau paintings to 1910 caused it to obtain a high creativity score, even above the cubism paintings from that time. On the Wikiart dataset, one of the highest-scored paintings was ``Tornado'' by contemporary artist Joe Goode, which was found to be mistakenly dated 1911 in Wikiart\footnote{\url{http://www.wikiart.org/en/joe-goode/tornado-1911} - accessed on Feb 28th, 2015 }.  A closer look at the artist biography revealed that he was born in 1937 and this painting was created in 1991\footnote{\url{ http://www.artnet.com/artists/joe-goode/tornado-9-2Y7erPME95YlkhFp7DRWlA2}}. It is not surprising for a painting that was created in 1991 to score very high in creativity if it was wrongly dated to 1911. These two example,  not only indicate that the algorithm works, but also show the potential of proposed algorithm in spotting wrong annotations in large datasets, which otherwise would require tremendous human effort.

\begin{figure*}[tp]
    \includegraphics[width=\linewidth]{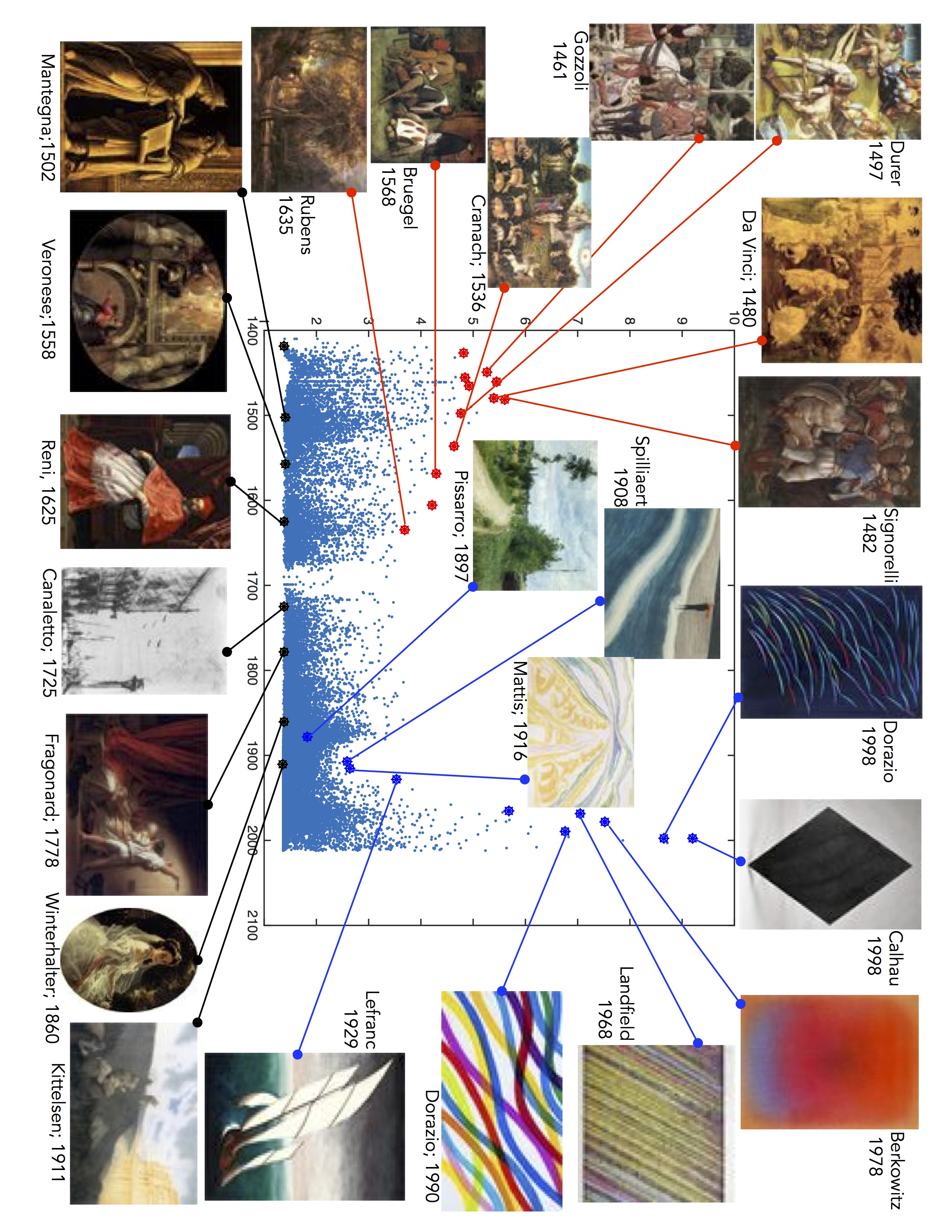}
    \caption{Creativity scores for 62K painting from the Wikiart dataset. The horizontal axis is the year the painting was created and the vertical axis is the scaled creativity score.}
  \label{F:WikiArt}
\end{figure*}

Figure~\ref{F:WikiArt} shows the creativity scores obtained for 62K paintings from the Wikiart datasets\footnote{For Figure~\ref{F:WikiArt} no temporal prior was used. We set K=500, $\alpha$=0.15.}.  Similar to the figures above, we plot the scores vs. the year of the painting.  We also randomly sampled points with low scores for visualization. The general trend in Figure~\ref{F:WikiArt} shows peaks in creativity around late 15th to early 16th century (the time of High Renaissance), the late 19th and early 20th centuries, and a significant increase in the second half of the 20th century.

\subsubsection*{Originality vs. Influence - Analysis of Religious Paintings}

In this experiment we investigate the effect of the two criteria of creativity: originality vs. influence. For this purpose we use the formulation in Eq~\ref{E:creativity_recursive2}.
In this experiment we used the religious paintings from the Wikiart dataset. This subset contains 5256 paintings in the period AD 1410-1993.  

\begin{figure*}[tp]
  \includegraphics[width=0.95\linewidth]{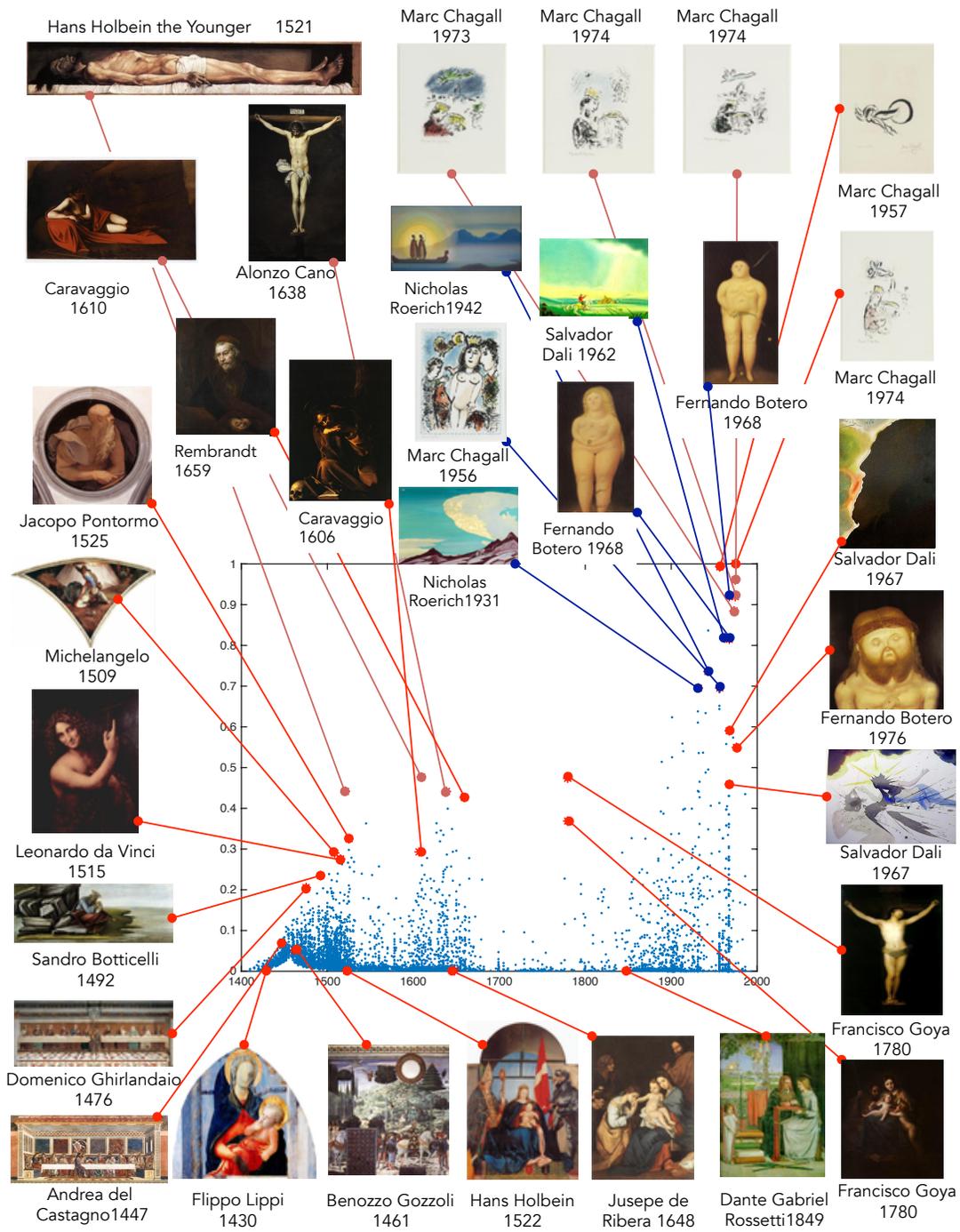}
    \caption{Creativity scores for 5256 religious paintings from the Wikiart dataset (AD 1410-1993), emphasizing originality in computing the creativity sores. The horizontal axis is the year the painting was created and the vertical axis is the scaled creativity score.}
  \label{F:WikiArtRel1}
\end{figure*}

\begin{figure*}[tp]
  \includegraphics[width=0.95\linewidth]{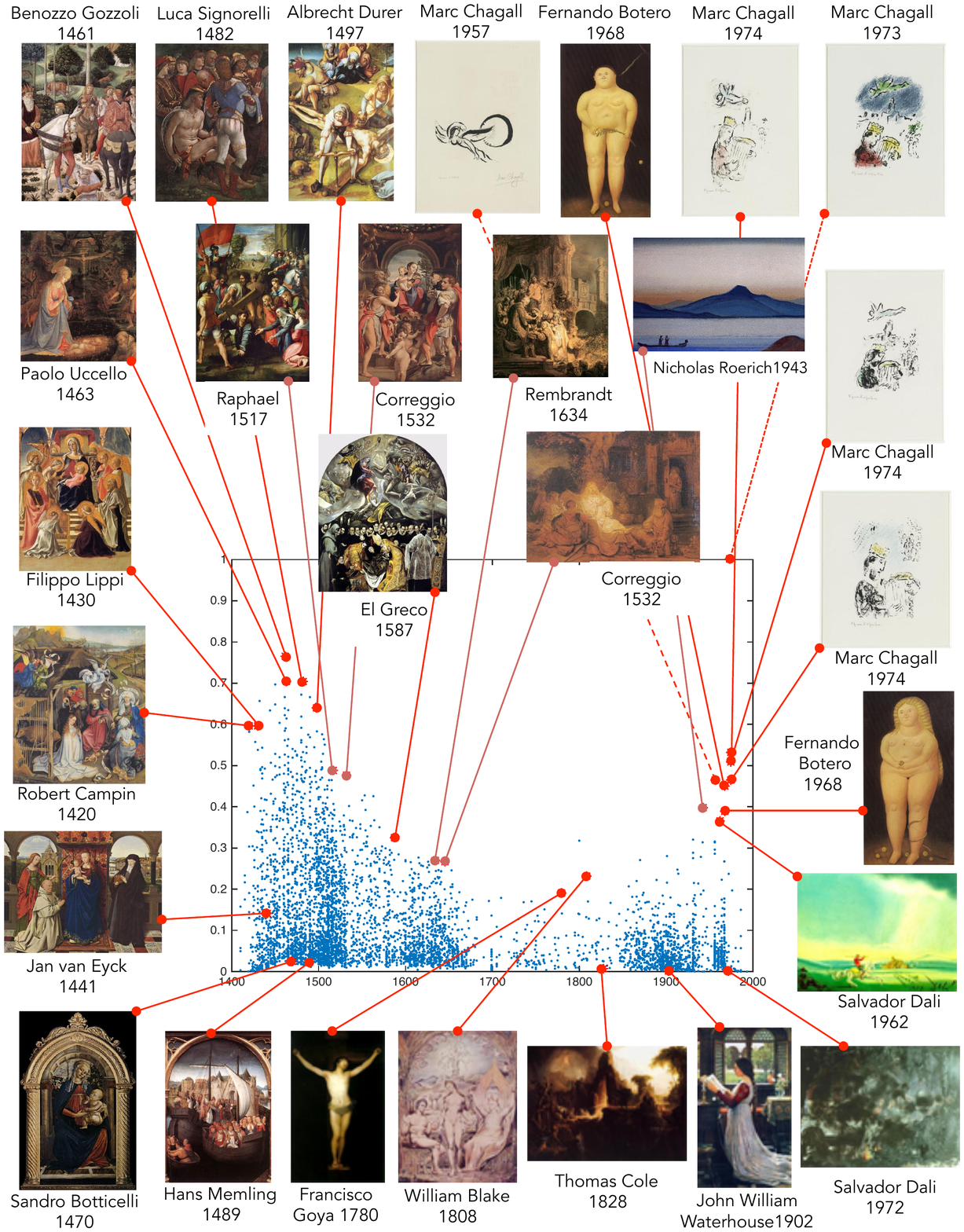}
    \caption{Creativity scores for 5256 religious paintings from the Wikiart dataset (AD 1410-1993), emphasizing influence in computing the creativity sores. The horizontal axis is the year the painting was created and the vertical axis is the scaled creativity score.}
  \label{F:WikiArtRel2}
\end{figure*}

Figure~\ref{F:WikiArtRel1} \&~\ref{F:WikiArtRel2} shows the creativity scores for this subset, where we set the parameter $\beta = 0.9$ to obtain the scores in Figure~\ref{F:WikiArtRel1} ( i.e., emphasizing originality) vs. setting the parameter $\beta = 0.1$ to obtain the scores in Figure~\ref{F:WikiArtRel2} (i.e., emphasizing influence). From the figures we can notice that emphasizing originality biases the scores towards modern paintings, while emphasizing influence biases the scores towards earlier paintings in the collection. Comparing the same painting in the two figures can contrast its novelty vs. its influence. For example, Francisco Goya's Crucified Christ (1780) scored very high in Figure~\ref{F:WikiArtRel1}, indicating its originality, and scored  lower in Figure~\ref{F:WikiArtRel2} when measuring its influence. However, in both cases that painting gets  higher creativity scores than other paintings from the same period. 

It is clear that emphasizing originality results in an monotonically increasing upper envelop in the plot at the period from 1400 until around 1520 (see Figure~\ref{F:WikiArtRel2}). This means that in this period there is a clear trend of increasing originality, where some paintings are pushing the upper envelop of the plot monotonically up. This up trend ends in the plot around 1520, which coincides with the end of the High Renaissance and the beginning of the Mannerism movement.  An interesting example of the paintings in the up trend of originality between 1400-1520 is Andrea del Castagno's 1447 Last Supper\footnote{A fresco located at the church of Sant'Apollonia in Florence} which is the earliest painting depicting the Last Supper in the analyzed  collection (see Figure~\ref{F:WikiArtRel1}). Domenico Ghirlandaio's 1476 last supper\footnote{A Fresco located in the abbey of San Michele Arcangelo a Passignano in Tavernelle Val di Pesa, near Florence}  scored higher along the upper envelop of the plot. In contrast, other versions of the Last Supper in the collection scored relatively lower, including Leonardo da Vinci's famous fresco.  Out of 18 paintings by da Vinci in this collection his St. John the Baptist (1515)  scored the highest (see Figure~\ref{F:WikiArtRel1}). In the modern era, some of the paintings that scored very high in this religious collection are by Marc Chagall, Fernando Boetro, Salvador Dali and Nicholas Roerich (see Figures~\ref{F:WikiArtRel1} and~\ref{F:WikiArtRel2} for details). 

%Compare to figure~\ref{F:WikiArtRel2}. 
%Compare to figure~\ref{F:WikiArtRel1}

\subsubsection*{Two-dimensional Creativity - Analysis of Portrait Paintings}

\begin{figure*}[tp]
  \includegraphics[width=0.95\linewidth]{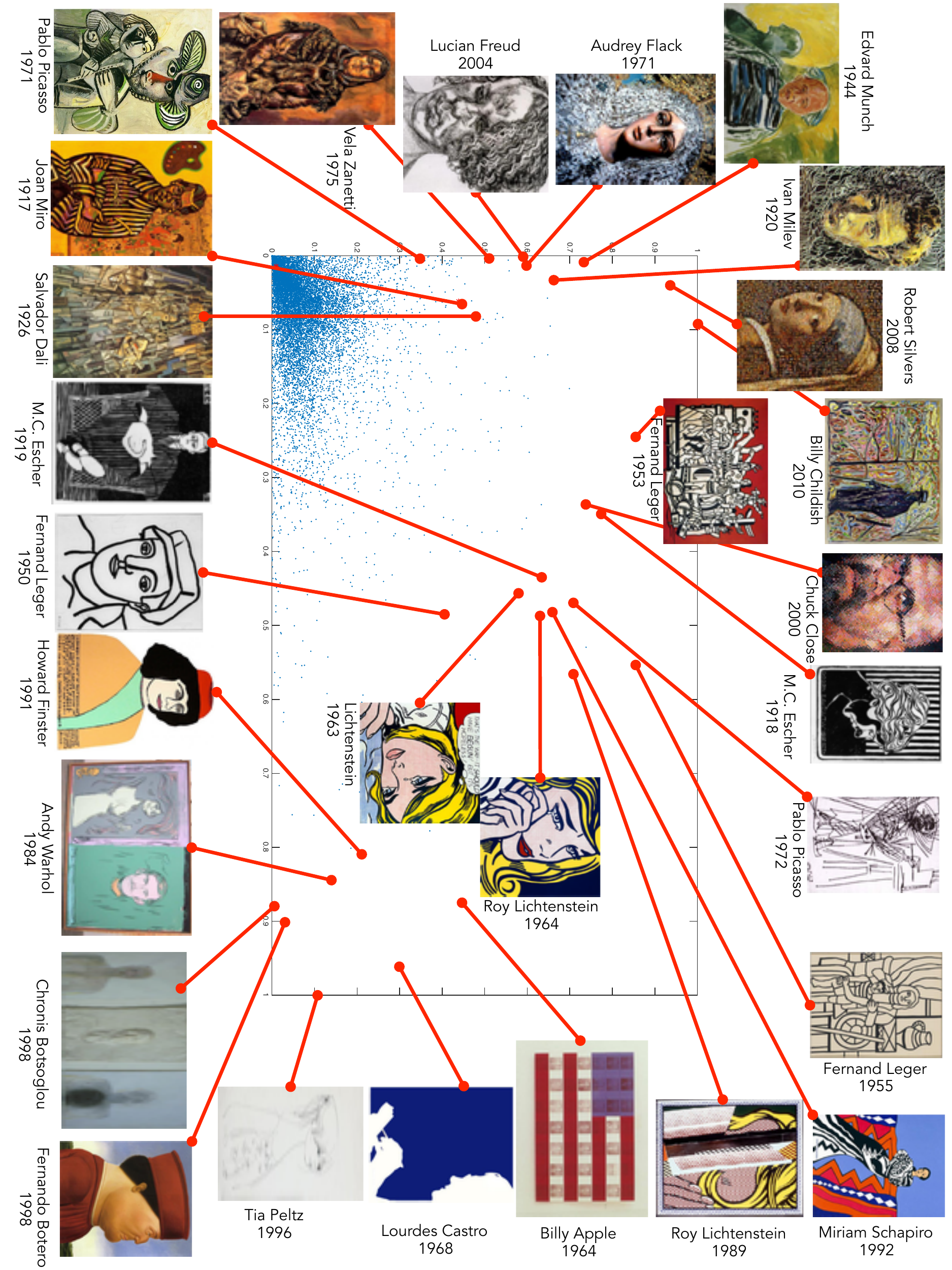}
    \caption{Two dimensional creativity scores for 12310 portrait paintings from the Wikiart dataset, ranging  from 1420 until 2011. }
  \label{F:WikiArtPortrait}
\end{figure*}

Figure~\ref{F:WikiArtPortrait} shows an example of two-dimensional analysis of creativity. In this experiment we used the subset of portrait paintings from the Wikiart dataset, which contains 12310 painting from the period AD 1420-2011. We analyzed creativity using the Classeme and GIST features as explained earlier, which yields two dimensions of creativity coordinates. Each point in the plot represents a single painting with two creativity scores. Unlike the previous figures, where we showed creativity vs. time, here we mainly show absolute creativity with respect to the two dimensions, i.e., we can not judge the relative creativity at any point of time from this plot. This makes the plot biased towards visualizing modern paintings.  It is clear from the plot that the horizontal axis correlates with abstraction in the shape and form, while the the vertical axis correlates with texture and pattern. 

\subsection{Time Machine Experiment}

\begin{table}[htdp]
\caption{Time Machine Experiment}
\label{T:TimeMachine}
\begin{center}
\small
\begin{tabular}{|l|l|l|}
\hline
Art movement & avg \% gain/loss &  \% increase \\
\hline
\multicolumn{3}{|c|}{Moving backward to AD 1600} \\
\hline
Neoclassicism &5.78\%$\pm$1.28& 97\%$\pm $4.8 \\
Romanticism & 7.52\%$\pm$ 2.04& 98\%$\pm$ 4.2  \\
Impressionism &14.66\%$\pm$ 2.78 & 99\%$\pm$3.2 \\
Post-Impressionism  &16.82\%$\pm$2.22& 99\%$\pm$3.1 \\
Symbolism &15.2\%$\pm$2.94  & 97\%$\pm$4.8\\
Expressionism & 16.83\%$\pm$2.43& 98$\%\pm$4.2 \\
Cubism & 13.36\%$\pm$2.43& 89\%$\pm$9.9\\
Surrealism &12.66\%$\pm$1.82& 95\%$\pm$7.1 \\
American Modernism &11.75\%$\pm$2.99& 84\%$\pm$8.4 \\
%Pop & 5.35\%$\pm$ 4.9& 72\%$\pm$1.23\\
\hline
\multicolumn{3}{|c|}{Wandering around to AD 1600} \\
\hline
Renaissance & 0.68 \%$\pm$ 2.05 & 39\%$\pm $5.7\\
Baroque &  2.85\%$\pm $ 1.09 & 71\%$\pm $19.7\\
\hline 
\multicolumn{3}{|c|}{Moving forward to AD 1900} \\
\hline
Renaissance & -8.13\%$\pm $ 2.02 & 20\%$\pm$10.5\\
Baroque & -10.2\%$\pm $2.03 & 0\%$\pm $0 \\
\hline
\end{tabular}
\end{center}
\label{default}
\end{table}%

Given the absence of ground truth for measuring creativity and the aforementioned wrong time annotations inspired us with a methodology  to quantitatively evaluate the framework. We designed what we call ``time machine'' experiment, where we change the date of an artwork to some point in the past or some point in the future, relative to its correct time of creation. Then we compute the creativity scores using the wrong date, by running the algorithm on the whole data. We then compute the gain (or loss) in the creativity score of that artwork compared to its score using correct dating. What should we expect from an algorithm that assigns creativity in a sensible way? moving a creative painting back in history would increase its creativity score, while moving a painting forward would decrease its creativity. Therefore, we tested three settings: I) Moving back to AD 1600: For styles that date after 1750, we set the test paintings back to a random date around 1600 using Normal distribution with mean 1600 and standard deviation 50 years (i.e.  $N(1600,50^2)$ ). II) Moving forward to AD 1900: For the Renaissance and Baroque styles, we set the test paintings to random dates around 1900 sampled from $N(1900,50^2)$. III) Wandering about AD 1600 (baseline): In this experiment, for the Renaissance and Baroque styles, we set the test paintings to random dates around 1600 sampled from $N(1600,50^2)$. 

Table~\ref{T:TimeMachine} shows the results of these experiments. We ran this experiment on the Artchive dataset with no temporal prior. In each run we randomly selected 10 test paintings of a given style and applied the corresponding move. We used 10 as a small percentage of the dataset (less than 1\%), not to disturb the global distribution of creativity. We repeated each experiment 10 times and reported the mean and standard deviations of the runs. For each style we computed the average gain/loss of creativity scores by the time move. We also computed the percentage of the test paintings whose scores have increased. From the table we clearly see that paintings from Impressionist, Post-Impressionist, Expressionist,  and Cubism movements  have significant gain in their creativity scores when moved back to 1600. In contrast, Neoclassicism paintings have the least gain, which makes sense, because Neoclassicism can be considered as revival to Renaissance. Romanticism paintings also have a low gain when moved back to 1600, which is justified because of the connection between Romanticism and Gothicism and Medievalism.
On the other hand, paintings from Renaissance and Baroque styles have loss in their scores when moved forward to 1900, while they did not change much in the wandering-around-1600 setting.

\section{Conclusion and Discusion}
The paper presented a computational framework to assess creativity among a set of products. We showed that, by constructing a creativity implication network, the problem reduces to a traditional network centrality problem. We realized the framework for the domain of visual art, where we used computer vision to quantify similarity between artworks. We validated the approach qualitatively and quantitively on two large datasets.  

The most important conclusion of this work is that,  when introduced with a large collection of paintings (and sculptures), the algorithm can successfully highlight paintings that is considered creative (original and influential). The algorithm achieved that without any knowledge about art or art history encoded in its input.  In most cases the results of the algorithm are pieces of art that art historians indeed highlight as innovative and influential. The algorithm achieved this assessment by visual analysis of paintings and considering their dates only.

  Besides this qualitative evidence, we also proposed a methodology for validating the results of the algorithm through what we denote as time machine experiments. This experiments quantitively validated the proposed algorithm. 
  
%Given the space limitation we could only show few results.

In this paper we focused on ``creative'' as an attribute of a product, in particular artistic products such as painting, where creativity of a painting is defined as the level of its originality and influence. However, the computational framework can be applied to other forms such as sculpture, literature, science etc. Quantifying creativity as an attribute of a product facilitates quantifying the creativity of the person who made that product, as a function over the creator's set of products.  Hence, our proposed framework also serves as a way to quantify creativity as an attribute for people.

%What artistic concept should be used to quantify creativity? is it the use of color, the use or perspective, subject matter, brush strokes, composition, etc? 

Clearly, it is not possible to judge creativity based on one specific aspect, e.g. use of color, perspective, subject matter, etc. For example it was the use of perspective that characterized the creativity at certain point of art history, however it is not the same aspect for other periods. This highly suggests the need to measure creativity along different dimensions separately where each dimension reflects certain visual aspects that quantify certain elements of art. The proposed framework can be used with multiple artistic concepts to achieve multi-dimensional creativity scoring.

\bibliographystyle{plainnat}
\bibliography{creativitybib}

\end{document}